\documentclass{article}

\usepackage{arxiv}

\usepackage{subcaption}
\usepackage{booktabs} 
\usepackage{graphicx}
\usepackage{amsmath,amsthm}
\usepackage{mathtools}
\usepackage{xspace}
\usepackage{epsfig}
\usepackage[vlined,linesnumbered,ruled,titlenotnumbered]
{algorithm2e}
\usepackage{color}
\usepackage{url}
\usepackage[graphicx]{realboxes}
\usepackage{tikz}
\usepackage{float}
\usepackage{booktabs}
\usepackage{makecell}
\usepackage{multicol}
\usepackage{multirow}
\usepackage{longtable}
\usepackage{array}
\usepackage{anyfontsize}
\usepackage{balance}
\usepackage{algpseudocode}
\usepackage{subcaption}
\usepackage{nicefrac}
\usepackage{microtype}      
\usepackage{caption}
\usepackage{rotating}

\newcommand{\ignore}[1]{}

\begin{document}

%\keywords{Evolutionary algorithms, open pit mine optimisation, open pit mine production scheduling, uncertainty, mine planning, staging}

\author{
   William Reid \\
   Maptek Pty. Ltd\\
   Adelaide, Australia\\ 
   \And
   Aneta Neumann \\
   Optimisation and Logistics\\ 
   School of Computer Science\\
   The University of Adelaide\\
   Adelaide, Australia \\
   \And
     Simon Ratcliffe \\
   Maptek Pty. Ltd\\
   Adelaide, Australia\\ 
   \And
     Frank Neumann \\
   Optimisation and Logistics\\ 
   School of Computer Science\\
   The University of Adelaide\\
   Adelaide, Australia \\
}

\title{Advanced Ore Mine Optimisation under Uncertainty Using Evolution}

\maketitle

\begin{abstract}
In this paper, we investigate the impact of uncertainty in advanced ore mine optimisation. We consider Maptek's software system Evolution which optimizes extraction sequences based on evolutionary computation techniques and quantify the uncertainty of the obtained solutions with respect to the ore deposit based on predictions obtained by ensembles of neural networks. Furthermore, we investigate the impact of staging on the obtained optimized solutions and discuss a wide range of components for this large scale stochastic optimisation problem which allow to mitigate the uncertainty in the ore deposit while maintaining high profitability.
\end{abstract}

\section{Introduction}\label{sec:introduction}

Evolutionary algorithms provide great flexibility in dealing with a wide range of optimisation problems. This includes highly constrained problems as well as problems involving dynamic and/or stochastic components. Their wide applicability has made evolutionary computing techniques popular optimisation techniques in areas such as engineering, finance, and supply chain management~\cite{DBLP:books/sp/2019DD}.

The area of mining where the goal is to extract ore in a cost efficient way poses large scale optimisation problems, and evolutionary computation techniques have successfully been applied in this area~\cite{DBLP:series/sci/BonyadiM16,DBLP:conf/cec/OsadaWBM13,DBLP:conf/gecco/SchellenbergLM16,DBLP:journals/jour/Ibrahimov2014}. 
We consider the problem of mine planning and focus on uncertainties which highly impact the mine planning process. Mine planning is one of the key optimisation problems in mining and a wide range of approaches have been developed over the years. The classical article of Lerchs and Grossmann~\cite{lerchs-grossmann} introduced the basic problem formulation and provided a dynamic programming approach. Over the years, a wide range of mine planning approaches taking different characteristics of this important real-world  optimisation problem into account have been studied in the literature. This includes integer programming approaches based on block scheduling~\cite{DBLP:journals/coap/MunozEGMQL18,DBLP:journals/ior/LetelierEGMM20} and heuristic techniques that are able to deal with various characteristics such as uncertainties of the problem~\cite{DBLP:journals/asc/GoodfellowD16,DBLP:journals/heuristics/MontielD17,DBLP:journals/cor/LamghariD20}. Different software products for carrying out mine planning and extraction sequences are available~\cite{maptek,minemax}.

We discuss the mine planning problem in the light of Maptek's mine planning optimisation software Evolution~\cite{maptek} which builds on evolutionary computing techniques. Evolution is representative of a new breed of commercially available algorithmic approaches to mine planning. One of the key features that set it apart from similar market offerings are its ability to handle large, complex data sets.
Algorithmic mine planning solutions are typically offered in two streams. The first being a linear programming approach that attempts to simplify complex models into linear relationships. The second utilise evolutionary algorithms, which can model non-linear relationships and arbitrarily complex constraints, where the results are considered near-optimal. Evolution is an example of the latter category, and is increasingly being used by large mining corporations globally for their life-of-mine extraction sequences.

We introduce and evaluate a new approach which allows the effect of uncertainties in the source geological data on which extraction sequences are based to be economically quantified. 
In mine planning the classical goal is to maximize net present value (NPV) over the life of the mine. This optimisation task involves deciding on whether to process given blocks of minerals based on the predicted amount of ore it contains. As the information on ore concentration can only be obtained by complex drilling processes the information about the grade of ore is highly uncertain for most of the material that needs to be considered.
In our approach, uncertainty quantification is done by an ensemble of neural networks that predict the grade of ore in the different blocks that should be processed. We have integrated our approach into the latest software release of Evolution.

A crucial part when running the Evolution software is the staging of blocks. This process divides the mine planning task into different stages that are processed sequentially. Given that the number of blocks is very large, usually in the hundreds of thousands, the staging process enables an efficient optimisation process by constraining the time at which a block can be processed. On the other hand, the staging heavily influences the quality of the solution to be obtained overall and at different periods in the extraction sequence. We exemplify this by showing different staging setups and the resulting profits that are obtained in the periods of a mine plan. Furthermore, staging has the potential to deal with uncertainties as it is possible to combine certain and uncertain areas of the mine into a stage which can lead to a reduction in the overall uncertainty in critical time periods of the life of a mine.
Our experimental investigations explore the uncertainty of typical mining schedules and compare the impact of different staging approaches on NPV and uncertainty in the different time periods of the life of a mine. 

The paper is structured as follows. In Section~\ref{sec:problem_formulation} we define the problem and describe the mineral resource block model.
Afterwards, we present Maptek's Evolution software in Section~\ref{sec:maptek} and introduce uncertainty qualification based on ensembles of neural networks in Section~\ref{sec:uncertain}. In Section~\ref{sec:staging}, we discuss the implications of staging on optimised solutions. We present our experimental results on uncertainty quantification and staging in Section~\ref{sec:experiments}. Finally, we discuss important open problems and research gaps, and finish with some concluding remarks.

\section{Problem Formulation}
\label{sec:problem_formulation}

The open-pit mine production scheduling problem, known as the open-pit mine block sequencing problem, is an important problem in open-pit mine planning as it determines the material that contributes to the sustainable utilization of mineral resources over the life of the mine~\cite{kennedy1990surface}. The overall goal is to find an optimal extraction sequence maximizing net present value (NPV).
Decisions on block scheduling in the open-pit mine production scheduling problem have to take into account several constraints~\cite{DBLP:conf/Johnson69,kennedy1990surface,DBLP:journals/anor/EspinozaGMN13}.
The set of constraints includes amongst others blending constraints, stockpile related constraints, logistic constraints, mill feed and mill capacity, reserve constraints and slope constraints.

In the following, we describe a basic model of the open pit scheduling problem similar to the one given in \cite{Morales2016} and it should be noted that the actual implementation might require the consideration of additional constraints and/or variations of the objective function.
A block model is given by a set of blocks $B$ and a set of destinations $D$.
Each block $i \in B$ has a certain number of parameters such as density, tonnage, ore grade, etc. These parameters permit to determine the economic value of every block $i \in B$ at a given time. 
We denote $T=\{1, \ldots, t_{\max}\}$ the set of time periods where $t_{\max}=|T|$ is the number of time periods. For each block $i \in B$ we assume a revenue of $r_{i}^t$ and a cost $c_i^t$ if block $i$ is sent at time $t$ for processing. On the other hand it encounters a mining cost $m_i^t$ if it is send to waste. 
In addition, the slope requirements for the set of blocks and other mining constraints are described by a set of precedence constraints $P$ as follows. The pair $(i,j)\in P$ describes a scenario where a block $i$ must be extracted by time $t$ if block $j$ needs to be extracted at time $t$. More precisely, $P$ is the set of precedence constraints, $i$ → $j$ if $(i, j) \in P$, block $i$ a has to be mined before block $j$. 

The value $(r^t_i$ - $c^t_i)$ is given for each block $i \in B$ at time $t \in T$ in the case if the block $i$ is sent to processing plant, and produces a cost $m^t_i$ if the block is sent to waste dump. We denote by $\tau_{i}$ the tonnage of block $i$.
For each period $t$ maximum limits $M^t$ on the amount of material that can be mined are imposed. Similar, for each period $t$ the maximum processing capacity $P^t$ i.e., the amount of ore that is milled have to be met. 

In our model we use two types of decision variables for each block $i \in B$ and time $t \in T$. The first type are the variables associated to the extraction for processing purposes for each block. A binary variable $x^t_i$ is one if block $i$ is extracted and processed in the period $t$ and zero otherwise. The second variable type describes the decision relating to the disposal of a block by sending it to the waste dump. The binary variable $w^t_i$ is one if block $i$ is extracted and sent to waste in the period $t$ and zero otherwise.

In this formulation, the objective function (\ref{eq:obj}) seeks to maximise the net present value of the solution (NPV) based on the a given discount rate $d$. Constraints (\ref{eq:con1}) ensures that is not possible to choose two different destinations for a block $i$ and that a block is only chosen at one point in time. 
Crucial constraints are \emph{the precedence constraints} (\ref{eq:con2}) which determine the extraction process of each block $i$ from surface down to the bottom of the ore deposit.
In order to provide for access and the stability of the pit walls it is not possible to mine a given block in a given time in the case that blocks in a defined pattern above have not already been extracted. 
The \emph{the mining constraints} (\ref{eq:con3}) ensure that the total weight of blocks mined during each period do not exceed the available extraction equipment capacity i.e., the mining capacity.
Similarly, the amount of material that can be processed in each period is restricted by the \emph{processing constraints} (\ref{eq:con4}).

The simplified mathematical model can be summarized as follows:
\begin{align}
	\mathbf{max} \quad   & NPV(x, w)= \sum_{t \in T} \left( \frac{1}{(1+d)^{t-1}} \sum_{i \in B} \left( (r^t_i - c^t_i)x^t_i - m^t_iw^t_i \right) \right)  \qquad \label{eq:obj}   \  \  \  \ 
	\\
	\mathbf{s.t.} \quad 
	& \sum_{t=1}^T x^t_i + w^t_i \leq 1 \qquad  \qquad  \qquad    \qquad  \qquad  \  \   \  \  \forall ~i~\in ~B \label{eq:con1}\\
	& (x^t_j +w^t_j) \leq \sum_{r=1}^t (x^r_i +w^r_i)  \qquad   \ \ \ \  \  \forall (i,j)\in P, t\in T \label{eq:con2}\\
		& \sum_{i \in B} \tau_{i}(x^t_i + w^t_i) \leq M^t  \qquad  \qquad  \qquad  \qquad   \  \  \  \  \ \forall  ~t\in T \label{eq:con3} \\
		& \sum_{i \in B} \tau_{i}x^t_i  \leq P^t  \qquad  \qquad   \qquad  \qquad  \qquad  \qquad \  \ \forall ~t\in T  \label{eq:con4}\\
& x, w \in \{0,1\}^{|B|\cdot|T|}
\label{con:binary}
\end{align}

There can be a number of additional supporting constraints, such as policies around stockpiling and product blend. Stockpiling is the process of storing mined material that will eventually go through the processing plant, but is deferred for mill capacity or economic reasons. Product blend constraints are used to ensure processed material meets a minimum grade, and can result in the schedule reflecting simultaneous mining from different areas of the pit, or using lower grade material from stockpiles to achieve a target grade. These have not been reflected above as they are not always required.

\subsection{The resource block model}

The mineral resource block model is a simplified representation of an ore body. The ore deposit is modelled in the form of rectangular, three-dimensional array of blocks that contain estimates of data such as dimensions, volume, spatial reference points, density of the material, grades of each block, the type of material, the shell and the corresponding pit stage, slope and bench. Additionally, the block contains a range of financial variables such as recovery cost, mining cost, processing cost, and rehabilitation costs. The information can often be reduced to the following: (1) ore grade, (2) ore tonnage and (3) destination (mill or waste dump) for each block. The set of blocks can be divided into two distinct subsets: the set of ore blocks sent to the mill and the set of waste blocks remaining. Waste blocks are not processed, but still need to be mined due the precedence constraints on ore blocks. Each block has an economic value at a given time period $t$ which represents the net present value that is associated with this particular block at time $t$ given by its revenues and associated processing cost. A waste block has negative financial value that is occurred by the cost of mining the block.

Blocks are grouped into a two layer hierarchy of stages and benches. Stages are assigned in the resource block model and methods for doing this are discussed. The benches within each stage are the sets of blocks with each horizontal layer of the block model.

Shells often form the basis for stages, and are calculated as subsets of blocks that can be mined at economic break-even over a range of likely ore prices. Shells are economically optimal subsets of the block model that honour precedence constraints. The shells can be generated using the well-known Lerchs-Grossmann algorithm~\cite{lerchs-grossmann}. The ultimate pit limit is the outer-most shell chosen based on the highest likely ore price and defines the total subset of blocks to be considered by a scheduling algorithm.

In the real world, the extraction of the ore based on sequentially mining Lerchs-Grossman shells is often impracticable. The mining shells are often discontinuous and consists of small number of blocks in between shells. Due to many complex mining processes which involve placement of machines and controlled blasting of rock, it is not profitable to mine small numbers of blocks in disconnected places. In order to tackle a real mining problem, we apply a stage design taking into account the representation of the valuable blocks.

\section{Maptek's Evolution Software}\label{sec:maptek}

Evolution is a suite of products offered by Maptek that provides scheduling engineers with tools to produce optimal extraction sequences constrained by many practical considerations. These products work together to collectively span multiple time horizons, ranging from:
\begin{itemize}
  \item life-of-mine: yearly scheduling showing returns for an entire mine's life.
  \item medium term: monthly scheduling used to optimise extraction sequence and equipment for a portion of the mine's life.
  \item short term: operational scheduling used to inform daily operations on required material volumes and grades, as well as assign given equipment to specific tasks.
\end{itemize}
For the purpose of our investigation, Maptek Evolution Strategy was used. Strategy is used for life-of-mine economic optimisation that aims to maximise net present value of the resource (NPV) by optimally ordering the stage/bench extraction sequence. It employs a dynamic cut-off grade policy for both the stockpiling and wasting of material.

Strategy's only required input is a block model as described. Along with the block model, Strategy requires yearly mining capacities and yearly plant capacities, as well as costs for mining, processing and selling the target ore. The optimisation weighs the cost of mining against the value of the ore being sold, dynamically choosing on a per stage per bench basis when and how to best handle material to maximise economic returns.

Strategy employs evolutionary algorithms to perform this optimisation. A dynamic cut-off grade policy introduces non-linearity into the optimisation and evolutionary algorithms require no compromises to handle this. The algorithm has been shown to perform robustly in acceptable time frames on very large whole-of-mine sized resource models. Another strength of an evolutionary algorithm approach is that the best solution is available at any given point in the run time. This means that a solution can be inspected for a given computation budget and released as 'good enough' without requiring guaranteed optimality.

Strategy is designed to always produce valid solutions. A \emph{valid} solution adheres to precedence and capacity constraints. Mutation and crossover operators are designed to preserve these constraints so provided seed genomes produce sequences that exhibit them, then the evolutionary algorithm will preserve them during selection and generation. A \emph{feasible} solution is not only valid, but also practical and economically viable to mine. This will typically require cash flows to remain positive after an initial start-up and with material movements aiming to consistently utilise mining equipment at or near full capacity.

A feasible solution isn't always the solution with the highest NPV for the resource model. This is because not all economic considerations are captured in this value. For example, solutions with periods of negative cash flow may make financing the operation infeasible even if this is for the benefit of larger future profits in net present terms. In this situation a solution that evens out revenues in each period will be considered feasible compared with another that may have a higher NPV.

Staging - the assignment of stages to divide the block model into sequentially mined subsets of blocks - is required by Evolution. It is used extensively in traditional manual scheduling but is also a useful tool to preserve feasibility whilst optimising for NPV. Staging is a discrete process and has a non-linear effect on NPV. Variability in the assigned domain of blocks and the ore grade within them also has a non-linear effect on the NPV of a given extraction sequence and this aspect is now able to be explored in the software from the work presented here.

\section{Uncertainty Quantification using neural networks}\label{sec:uncertain}

A mineral reserve block model is inherently uncertain because the estimations of ore grade in each block are derived from samples obtained by drilling into the rock and the spacing of drilling is typically much wider than the block resolution. Many interpolation methods exist for combining samples in the local neighbourhood of a block to estimate its grade value. Those with a basis in geological processes are favoured. These methods typically involve a categorisation of the rock mass into \emph{lithological domains}. Such domains typically delineate separate mineralisation processes that have occurred under different physical conditions, such that the formation of ore and its concentration in one domain is considered independent to that in another. The domain classification then guides the choice of samples and the weight to give to them in a geostatistical interpolation method such as Kriging~\cite{krige1951statistical} or inverse distance weighted averaging~\cite{shepard1968two} to derive an estimate of the grade for each block in the block model. 

Quantifying the uncertainty of any estimation process involving domains is complicated to do analytically because complex 3D geometries are often involved with the interfaces between domains~\cite{DBLP:journals/jilsa/DuttaBGM10,Nezamolhosseini2017}. Conditional simulation is a commonly used technique in geostatistics to quantify the geological uncertainty in a multi-domain reserve block model~\cite{Ortiz2004AMA}. This method can be thought of as a sensitivity analysis of grade value estimations and their distribution throughout the model based on varying one or several of the parameters involved in one or several geostatistical interpolation methods. The range through which to vary the parameters and the methods to include in the simulation are all subject to qualitative choices and are difficult to approach for a novice in the field.

Here we employ a simpler technique to quantify the locations and ranges of uncertainty which has been introduced in~\cite{Sullivan2019}. The method produces data that is adequate to illustrate the economic sensitivity of different extraction sequences based on the spatially varying uncertainty that arises from a population of different interpolations of the input sample data. The method  simultaneously allows variation in lithological domain boundaries and grade distributions within the domains to be explored. The approach in summary is to train a deep neural network to fit the input sample data in its three spatial dimensions, an arbitrary number of continuous dimensions corresponding to grades of elements of interest and a categorical dimension corresponding to the domain. The fit is performed from random starting weights and employs standard deep learning model fitting techniques based on gradient descent to minimise a differentiable error function until the goodness-of-fit is within a specified tolerance. By exploiting the property of such networks to converge on a set of weights from different random initialisations, we compute an ensemble of trained networks that statistically interpolate the input data equally well.

Maptek's DomainMCF machine learning geological modelling software was used to produce an ensemble of ten models based on this technique. These models are then evaluated into each block in a block model to produce a population of grade and domain estimates for each block. These ensembles are the basis for quantifying the economic uncertainty of optimised extraction sequences.

\section{Staging}
\label{sec:staging}
Staging is an important component that serves as an input to Evolution. 
It divides the optimisation problem into stages that are processed sequentially. Staging is therefore a partitioning of the given set of blocks $B$ into stages $S_1, \ldots, S_k$ where each block in stage $S_i$ is processed before each block in stage $P_j$ iff $i<j$.  For the given partitioning $S_i \cap S_j= \emptyset$ iff $i \not =j$ and $\cup_{i=1}^k S_i=B$ is required. In practice, the requirement of processing a stage completely before starting the next one is not strict and there might be some time overlaps in processing different stages.
Note that a trivial staging is obtained by using only one stage $S_1$ and setting $S_1=B$ which implies that all blocks belong to the same stage. The drawback of this is that all blocks have to be considered at the same time which does not break down the optimisation problem. Furthermore, benches are scheduled completely in a single stage and breaking benches down by dividing them into different stages can lead to better overall solutions.

Several methods have been proposed that do not use staging in mine optimisation and schedule blocks directly. These are collectively known as direct block scheduling methods. They remain an active area of research because they still have not solved the benefits afforded by staging methods around some aspects of practicality even though they may yield an extraction sequence of higher NPV. For example, staging methods inherently create a controlled number of active mining areas and so afford a direct means of localising where equipment needs to be and where it needs to go next on relatively long time frames. This enables roads to be constructed in time and without interfering with other passages, distances over which equipment needs to be moved to be minimised and other similarly advantageous considerations for practical mining. Direct block scheduling methods require explicit constraints to gain these advantages and without them will often result in extraction sequences that involve mining small pockets of material from impractically distant locations.

Staging helps to take a problem from hundreds of millions of variables into a few hundred that can be optimised with complex non-linear relationships. It introduces practical constraints on the order of blocks in a schedule. Strategy uses stages and considers combinations of stage and bench. A stage and bench combination is all the blocks that share the same stage and position on the Z axis (the bench).
When considering what blocks to mine in each period, the optimisation process will sum the values of each block in a stage and bench and mine it in one large chunk. Without staging, Strategy would be forced to mine the entirety of each bench before moving on. This will result in significant mining of waste before any ore could be reached. In effect, staging allows high value ore to be targeted early on in the process, leaving the movement of waste from other stages to later in the process.

The creation of these stages are used as a loose guide for a logical mining sequence. Typically, stage $1$ will be mined before stage $2$, but there will be some overlap. As mining hits ore in stage $1$, the start of stage $2$ - which will often be removal of waste blocks to get to future ore blocks - would begin. This simultaneous mining of ore and waste in stages ensures a consistent positive cash flow.
Through the process of staging, a number of practical mining rules are adhered to. There is an assumption that a completed stage design will follow some basic mining practices. These are - but not limited to - the rules outlined in Section~\ref{sec:maptek} which Strategy uses to produce valid mining solutions.

\section{Experimental Investigations}\label{sec:experiments}
 Our approach on uncertainty quantification has been implemented into Evolution and is part of the latest software release. We now investigate the uncertainty of optimised solutions introduced by the neural network approach. We interpret the solutions obtained from a mine planning perspective in terms of economic risks associated with the obtained solutions. Furthermore, we investigate optimised solutions for different staging approaches and show that this leads to significantly different results.
 
\begin{figure}[t]
\centerline{\includegraphics[width=\textwidth/2]{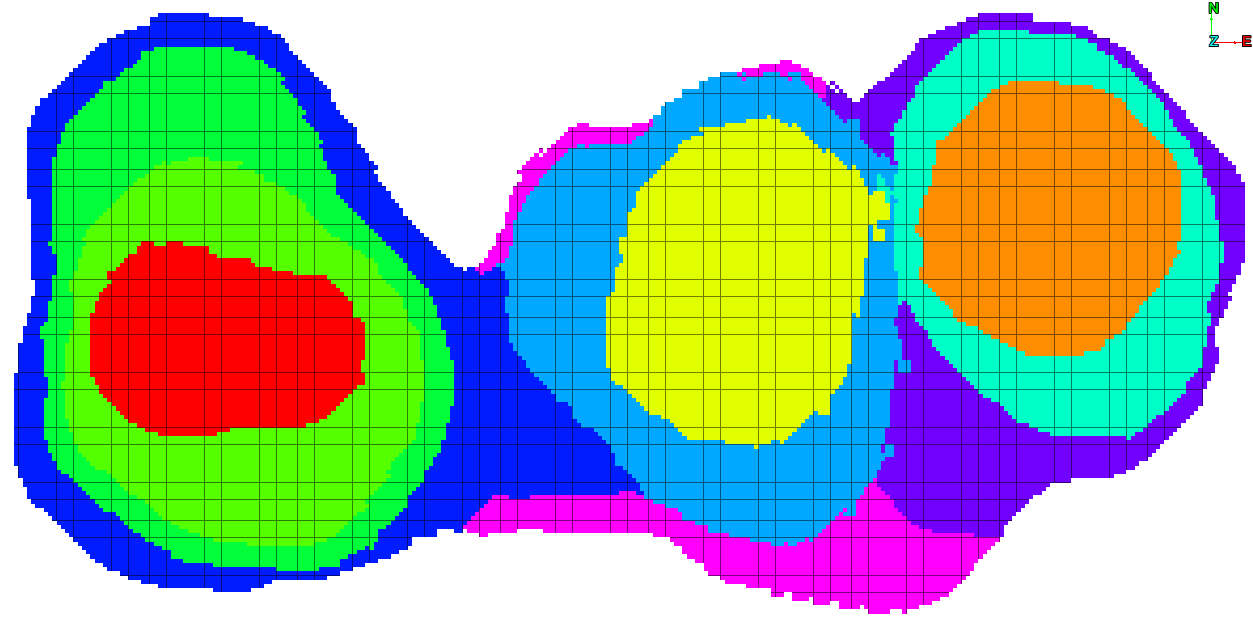}}
\caption{Example of model staging. This model has 10 total stages, created from 21 individual shells.}
\label{fig:model_staging}
\end{figure}
\begin{figure}[t]
\centerline{\includegraphics[width=\textwidth/2]{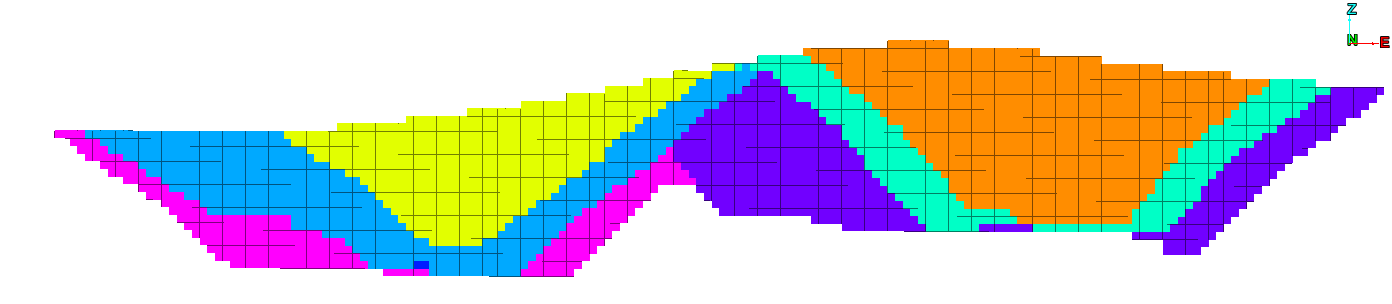}}
\caption{A cut along the Y axis of the model showing the eastern section of the pit.}
\label{fig:model_staging_cut}
\end{figure}

\subsection{Problem setup}
Using the same input drilling data we generated 10 individual block models using the neural network technique. We will call this collection of models an ensemble, with each individual model being a member of the ensemble.

We then combined the ensemble into a single model, known as the aggregate model. This aggregate model took the median domain between the ensemble members, and the mean grade for the selected domain. We used this aggregate to create an ultimate pit using Lerchs-Grossmann~\cite{lerchs-grossmann} pit optimisation. From here, we could begin to stage the model. 

During testing we recognised a number issues when comparing results from different staging techniques. The largest being that in order to compare the staging between tests, constraints manually applied to positively drive feasibility in one should not then negatively drive feasibility in another. Typically, a mining engineer would manually apply constraints of this nature such as:
\begin{itemize}
    \item Bench turnover: The number of bench's to be mined per period
    \item Stage availability: Holding a stage back from the optimiser in order to achieve some aspects of feasibility 
\end{itemize}
and many more. The application of these constraints is very specific to each individual staging and so could not sensibly remain constant across all our tests. As such, no such additional constraints were placed on the optimisation. Instead, each test shared the same `calendar'. A calendar sets maximum tonnages for mining capacity, plant processing and also economics such as ore price, processing and mining costs, and other constants on a per-period basis.

For our investigations we used representative mining values for the copper ore under consideration; a processing cost of $\$15$ AUD per tonne, and a mining cost of $\$4.20$ AUD per tonne. Mining capacity was set to $25$ million tonnes per period, with periods defined in yearly increments. The resulting block model had $190\,000$ blocks, and given the significant over-burden meant the processing plant was only required from the middle of period $3$. Plant capacity began at $5$ million tonnes per period, before being upgrade to $9$ million tonnes in period $9$. There were no constraints placed on stockpiling capacity. Included in the economics was a minimum cut-off grade of $0.25\%$ copper, and a price per tonne of $\$7\,673$ AUD. Selling cost (the administrative cost involved in selling the ore) was set at $\$1$ AUD per tonne, and rehabilitation cost (the cost to rehabilitate the site at the end of the mine life) was also set at $\$1$ AUD per tonne. The deposit included a number of different copper rich domains. These domains ranged in copper recovery - the amount of material that can realistically be recovered by milling - from $75\%$ to $92\%$.
To begin our tests, we used the individual shells generated by the Lerchs-Grossmann pit optimisation. These shells guided subsequent staging approaches. Figure~\ref{fig:model_staging} and~\ref{fig:model_staging_cut} show an example of a staged model.

As part of the optimisation process, the Strategy engine picks a near-optimal per-period stage/bench extraction sequence for the aggregate model and its associated staging. Once optimisation is complete, the engine takes each individual model and replays the same stage/bench sequence, evaluating the per-period economics. This results in us getting back $11$ different evaluations; $1$ for the aggregate model, and $10$ in total for the ensemble. This allows us to plot the economics for each member of the ensemble on a per-period basis for comparison.

The block properties that vary due to geological uncertainty are domain and grade. The assigned domain influences the processing method and associated cost this may result in a decision to waste a block instead of process it. Changes in grade can force an ore block to now sit below cut-off, thus classifying it as waste. Re-classifications like this are thought to be where uncertainty will have the biggest economic impact.

\subsection{Staging approaches}
With the problem now consistently formulated independently of staging, we started to consider how stage design could affect the resulting charts. With this in mind we opted for $4$ different stage designs, each with $6$ individual stages. Note that each staging mechanism is practical and potentially feasible.
\paragraph{Lazy staging}
This stage design involved aggregating sequential Lerchs-Grossmann shells into stages. This represents an NPV-maximising stage design without any consideration of geological uncertainty.

\begin{figure}[]
\begin{subfigure}[b]{8cm } 
\centering
\includegraphics[width=\textwidth]{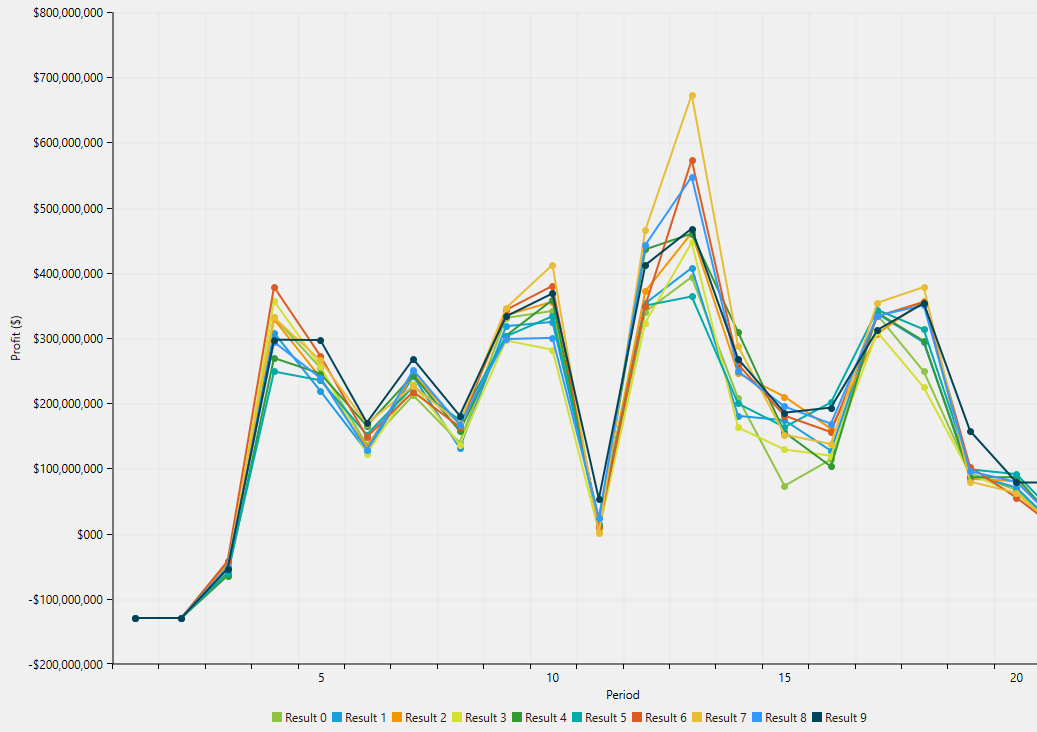}
\caption{Lazy staging}
\label{fig:profit_a}
\end{subfigure}
\begin{subfigure}[b]{8cm } 
\centering
\includegraphics[width=8.0cm,height=5.65cm]{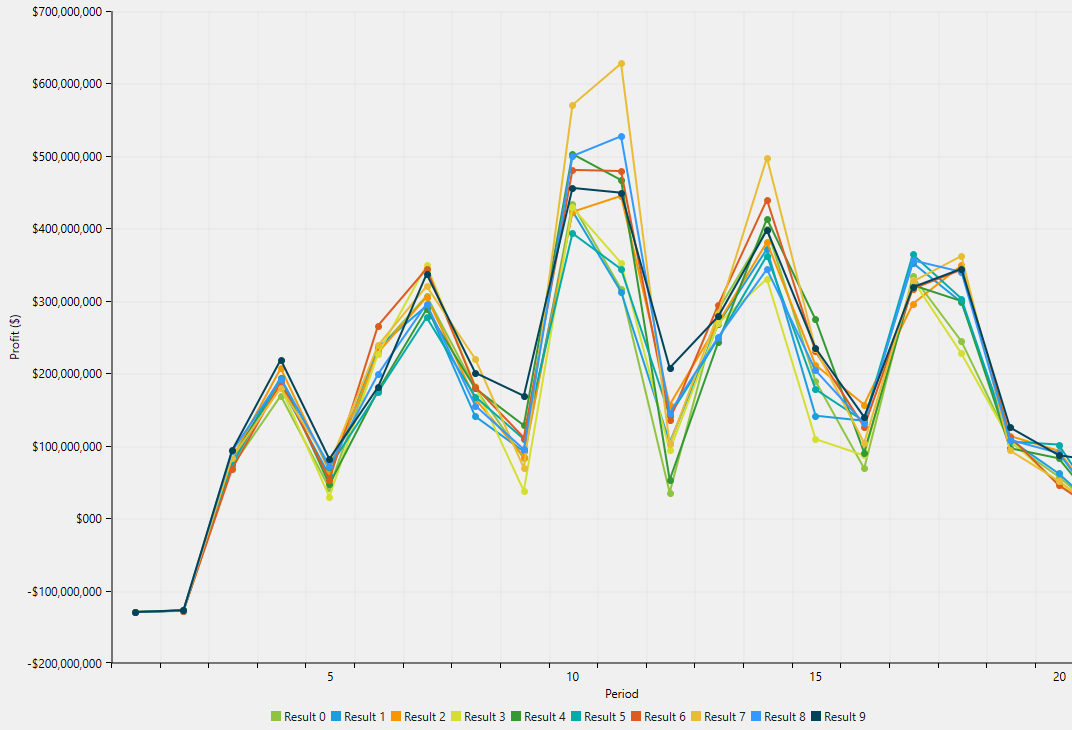}
\caption{Expected staging}
\label{fig:profit_b}
\end{subfigure}
\begin{subfigure}[b]{8cm}
\centering
\includegraphics[width=\textwidth]{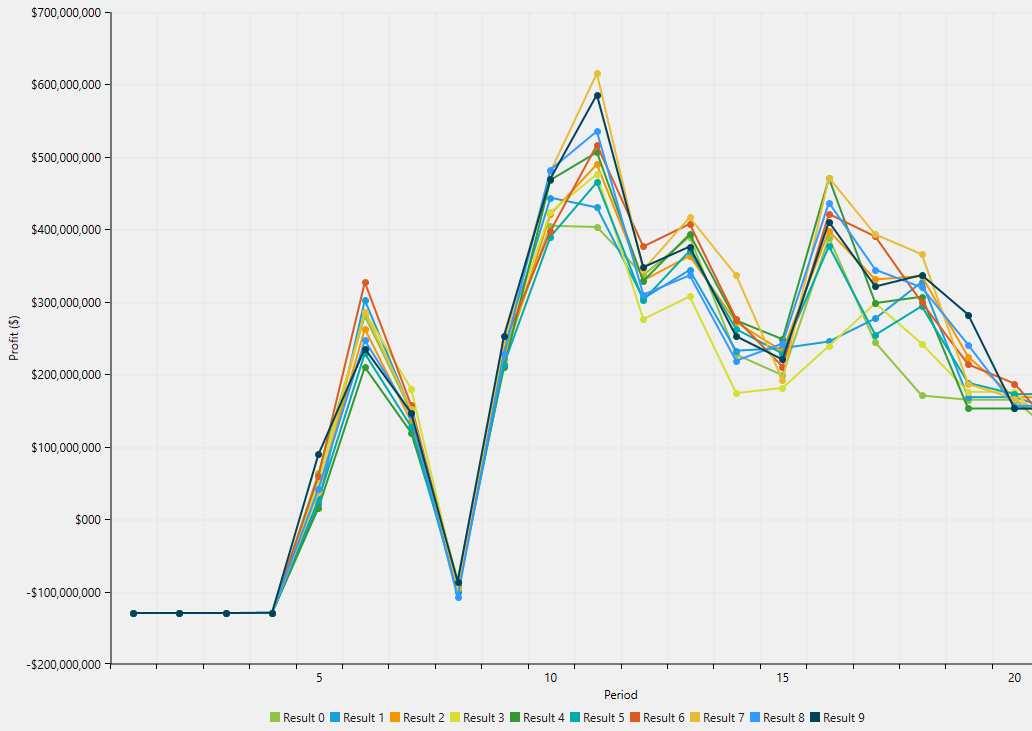}
\caption{Worst case staging}
\label{fig:profit_c}
\end{subfigure}
\hspace{0.4cm}
\begin{subfigure}[b]{8cm}
\centering
\includegraphics[width=\textwidth]{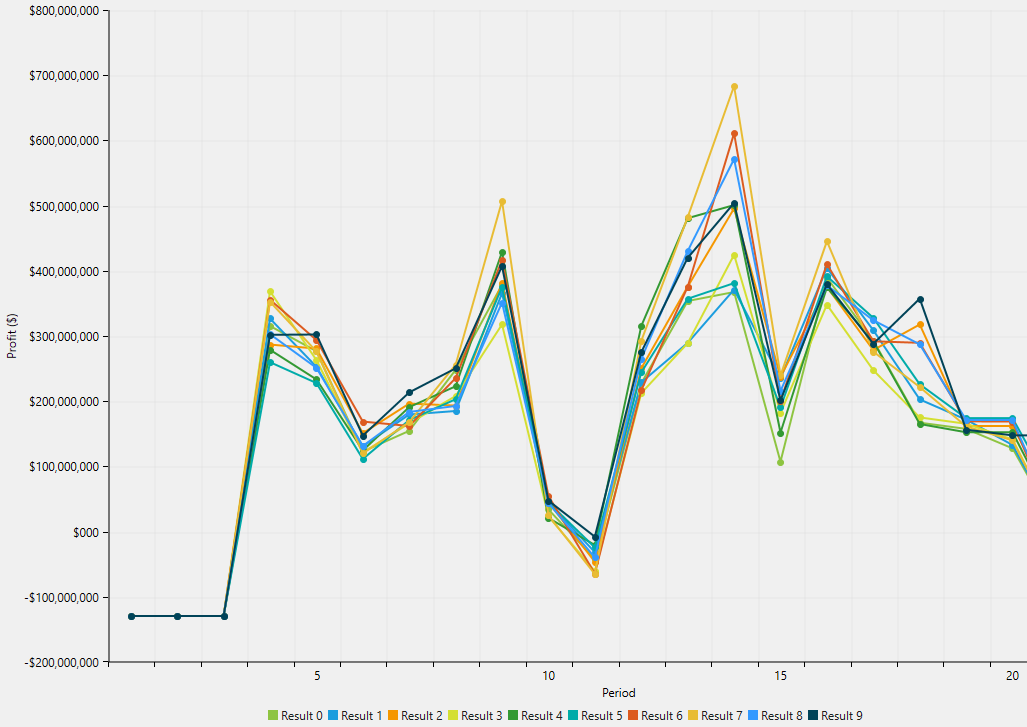}
\caption{Levelled uncertainty staging}
\label{fig:profit_d}
\end{subfigure}
\caption{Profit per period per ensemble member for four different staging approaches.}
\label{fig:profit}
\vspace{-0.5cm}
\end{figure}

\begin{figure}[t]
\begin{subfigure}[b]{8cm}
\centering
\includegraphics[width=\textwidth]{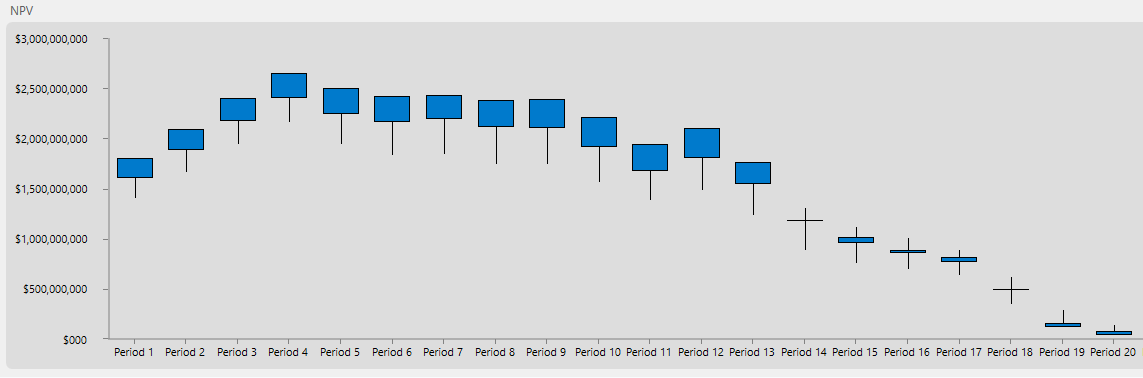}
\caption{Lazy staging}
\label{fig:npv_a}
\end{subfigure}
\begin{subfigure}[b]{8cm}
\centering
\includegraphics[width=8.0cm,height=2.65cm]{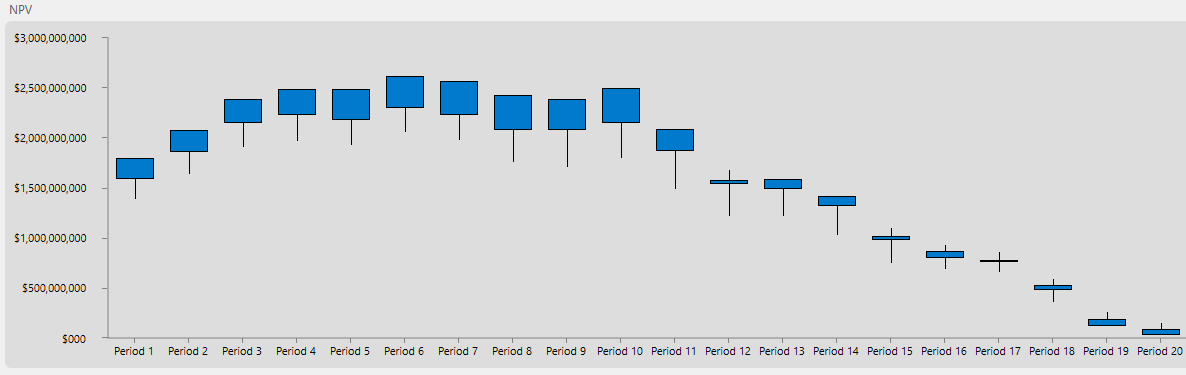}
\caption{Expected staging}
\label{fig:npv_b}
\end{subfigure}
\begin{subfigure}[b]{8cm}
\centering
\includegraphics[width=\textwidth]{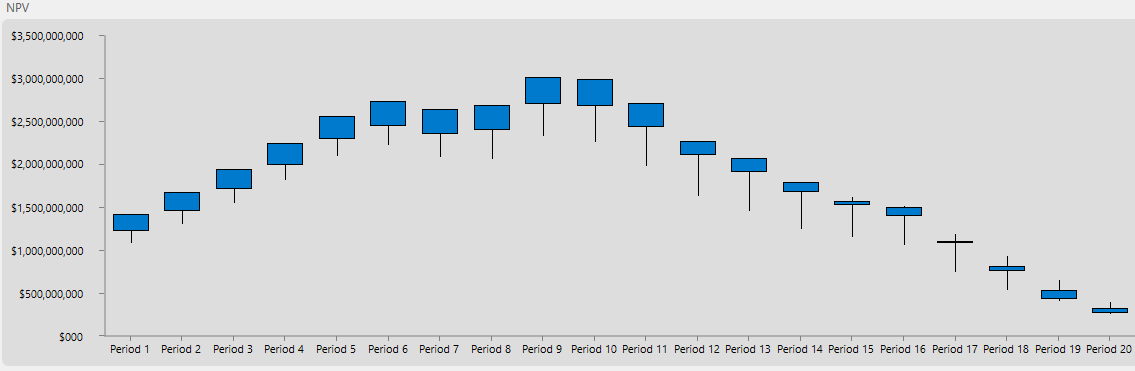}
\caption{Worst case staging}
\label{fig:npv_c}
\end{subfigure}
\hspace{0.4cm}
\begin{subfigure}[b]{8cm}
\hspace{1cm}
\centering
\includegraphics[width=\textwidth]{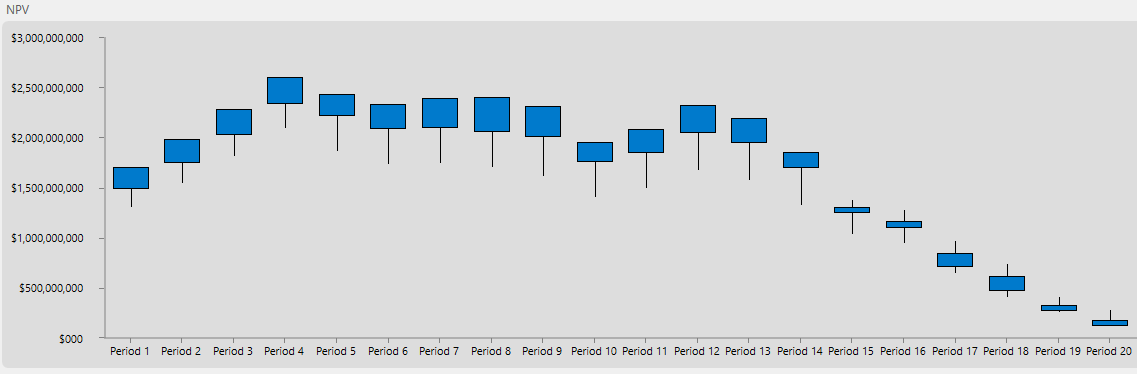}
\caption{Levelled uncertainty staging}
\label{fig:npv_d}
\end{subfigure}
\caption{Box and whisker plot showing the remaining resource NPV per period for four different staging approaches.}
\label{fig:whisker}
\vspace{-0.5cm}
\end{figure}

\paragraph{Expected staging}
For this stage design, a professional mining engineer with knowledge of the deposit staged the model based on a number of factors: maximising NPV whilst keeping the plant full and maintaining a positive cash flow across all periods, again without consideration of geological uncertainty.

\paragraph{Worst case staging}
To best try and see an effect from uncertainty, this staging isolates all ore blocks with a grade standard deviation greater than $1$\% to only two stages. The remaining $4$ stages consisted the ore blocks with a grade standard deviation of less than or equal to $1$\% and the waste blocks.

\paragraph{Levelled uncertainty staging}
With the results of the last three stage designs in mind, the last attempt was to try and level out uncertainty between the $6$ stages. By grouping pockets of uncertainty together with certain ore, the hope was the combination would result in a reduction in overall per-period uncertainty. The levelling was done manually based on visual inspection of grade variance.

\subsection{Experimental results}
For our investigations, we refer to average NPV to represent the models feasibility, and profit range to represent what effect uncertainty has on the overall schedule. 
The results of four different staging approaches with respect to per period profit are shown in Figure~\ref{fig:profit} and the NPV whisker plots are shown in Figure~\ref{fig:whisker}. The figures show screenshots of our approach implemented into Evolution. 
The overall summary of the results in terms of the average NPV and the total profit range (difference of highest total profit and lowest total profit predicted by the ensemble) is as follows.

\begin{itemize}
\item Lazy staging: average NPV of $\$1.585$B, total profit range of $\$1.84$B.
\item Expected staging: average NPV of $\$1.569$B, total profit range of $\$1.929$B.
\item Worst case: average NPV of $\$1.206$B, total profit range of $\$1.98$B.
\item Levelled uncertainty staging: average NPV of $\$1.485$B, total profit range of $\$1.99$B.
\end{itemize}

Figure~\ref{fig:profit} shows the profit results of $4$ different stage designs for $20$ periods. The results for each of the $10$ ensemble members are shown as colored lines. The spread of the lines shows the range of economic uncertainty introduced by the geological uncertainty in each period and comparison between the charts highlight how staging can mitigate or exacerbate this. Additional details on the minimum, maximum, average and standard deviation of profit values per time period are given in Table~\ref{tab:results}. 

In Figure~\ref{fig:profit} (\subref{fig:profit_a}) we see the results in terms of the profit achieved at each period for lazy staging approach for $10$ models. In this setting, the obtained function values for period $t = 4, 10, 13$ are the highest ($\$377M, \$411M, \$673M$), respectively, and the lowest profit value of $-\$0.5M$ is obtained in the period $t = 11$. The obtained uncertainty variance is the highest for $t = 13$. Note that in practical mining scheduling often the profit at the beginning of mining is negative due to mining of covering waste blocks.
Figure~\ref{fig:profit} (\subref{fig:profit_b}) shows the results for the expected staging approach. The obtained profit is highest ($\$570M, \$627M, \$497M$) in the period $t = 10, 11, 14$, respectively. In contrast to the previous approach, we observe that the highest uncertainty variance is obtained for period $t = 11$.
Figure ~\ref{fig:profit} (\subref{fig:profit_c}) shows the worst case staging approach generates the worst profit values overall and obtains the highest profit values ($\$481M, \$615M, \$471M$) for period $t = 10, 11, 15$, respectively. More surprising, not only is the uncertainty variance highest for $t = 11$ but we also observe high uncertainty at the end of the life of the mine, namely in the period $t = 16, 17, 18, 19$.
Finally, Figure~\ref{fig:profit} (\subref{fig:profit_d}) shows the profit obtained using levelled uncertainty staging approach. In this scenario, where geological uncertainty was considered in the stage design, the highest profit $\$683M$, occurred later than in previous approaches in the period $t = 14$ and the lowest negative profit $\$-66M$, in the period $t = 11$ except the first three periods. The obtained uncertainty variance shows the similar course as in the Figure~\ref{fig:profit} (\subref{fig:profit_a}) i.e., the highest variance is observed in the period $t = 14$ where also the highest profit occurred.

\begin{sidewaystable}[t]

\caption{\label{tab:}Maximum (\text{max}), minimum (\text{min}), mean (\text{mean}), and standard deviation (\text{std}) in terms of profit (\$) for four staging approaches. 
\label{tab:profit_results}
}

\begin{scriptsize}

\setlength{\tabcolsep}{2.3pt} 
\renewcommand{\arraystretch}{1.6} 

\begin{tabular}{rrrrrrrrrrrrrrrrr}
\toprule
\multicolumn{17}{c}{\textbf{P R O F I T}}\\
\cmidrule(l{3pt}r{3pt}){2-17}

                      & \multicolumn{4}{c}{\textbf{Lazy staging (1)}}                                                                         & \multicolumn{4}{c}{\textbf{Expected staging (2)}}                                                                         & \multicolumn{4}{c}{\textbf{Worse case staging (3)} }                                                                         & \multicolumn{4}{c}{\textbf{Levelled uncertainty staging (4)} }                                                                         \\
\cmidrule(l{3pt}r{3pt}){2-5} \cmidrule(l{3pt}r{3pt}){6-9}
\cmidrule(l{3pt}r{3pt}){10-13} \cmidrule(l{3pt}r{3pt}){14-17}

\multicolumn{1}{c}{\textbf{t}} & \multicolumn{1}{c}{\textbf{max}} & \multicolumn{1}{c}{\textbf{min}} & \multicolumn{1}{c}{\textbf{mean}} & \multicolumn{1}{c}{\textbf{std}} & \multicolumn{1}{c}{\textbf{max}} & \multicolumn{1}{c}{\textbf{min}} & \multicolumn{1}{c}{\textbf{mean}} & \multicolumn{1}{c}{\textbf{std}} & \multicolumn{1}{c}{\textbf{max}} & \multicolumn{1}{c}{\textbf{min}} & \multicolumn{1}{c}{\textbf{mean}} & \multicolumn{1}{c}{\textbf{std}} & \multicolumn{1}{c}{\textbf{max}}& \multicolumn{1}{c}{\textbf{min}} & \multicolumn{1}{c}{\textbf{mean}} & \multicolumn{1}{c}{\textbf{std}} \\ \hline
1                     & -130250000              & -130250000              & -130250000               & 0.0                       & -130250000              & -130250000              & -130250000               & 0.0                       & -130250000              & -130250000              & -130250000               & 0.0                       & -130250000              & -130250000              & -130250000               & 0.0                       \\
2                     & -130250000              & -130250000              & -130250000               & 0.0                       & -127686754              & -128559124              & -128076490               & 243682.6              & -130250000              & -130250000              & -130250000               & 0.0                       & -130250000              & -130250000              & -130250000               & 0.0                       \\
3                     & -43003516               & -65238038               & -53494867                & 7500500.0              & 93001540.2              & 67567509.4              & 80687311.1               & 8226357.5              & -130250000              & -130250000              & -130250000               & 0.0                       & -130250000              & -130250000              & -130250000               & 0.0                       \\
4                     & 377666413               & 247890113               & 313721272                & 39142231.9              & 218149045               & 168226841               & 189805079                & 14313557.9              & -129794911              & -130240649              & -130145265               & 144482.5              & 367832346               & 259189975               & 314227365                & 35783141.5              \\
5                     & 296127619               & 218012252               & 251672750                & 22165891.9              & 81011752.2              & 28819070.2              & 58600949.4               & 16406513.6              & 88693806.8              & 14915434.3              & 43601463.3               & 22286506                & 301647149               & 227163533               & 265172607                & 24613965.7              \\
\cmidrule{1-17}
6                     & 169539650               & 121938844               & 144194954                & 18338564.4              & 265053007               & 173179102               & 215034038                & 31266553.1              & 326229459               & 209235340               & 265896134                & 36358016.1              & 168098131               & 110723541               & 132181782                & 17534232.2              \\
7                     & 267020158               & 211454610               & 235159816                & 16766315.6              & 348489523               & 276429252               & 311037144                & 24907894.7              & 179146701               & 118519573               & 145160118                & 16673731.6              & 212831164               & 153862905               & 177876120                & 17947427.4              \\
8                     & 180310851               & 130687008               & 158126583                & 17171760.4              & 219158586               & 139847250               & 176464398                & 22352769.8              & -84961879               & -108577862              & -94198476                & 6891740.6             & 255338344               & 184280502               & 218793008                & 26598563.9              \\
9                     & 345554357               & 295667982               & 320060358                & 19276574                & 168164965               & 36817076.7              & 97146590.4               & 35213116.4              & 252598223               & 208935015               & 230983938                & 13853225.2              & 506256332               & 317132886               & 392349639                & 51441530.5              \\
10                    & 411434723               & 281275037               & 344964781                & 38143933.6              & 569627084               & 392546645               & 460447937                & 52548952.9              & 481754826               & 389184095               & 437471124                & 35326818.5              & 53862053.2              & 20211591.9              & 37589529.2               & 11688540.8              \\
\cmidrule{1-17}
11                    & 52438199.1              & -486423.6              & 15677302                 & 14878462.7              & 626869693               & 311816368               & 431274119                & 101642852               & 614589821               & 402742016               & 501763759                & 65218777.5              & -8531588.1              & -65753828               & -40946358                & 19497027.5              \\
12                    & 464568298               & 322983719               & 383684339                & 50020521.8              & 206731579               & 34644789.6              & 116739055                & 50845326.8              & 375660147               & 275456244               & 324942187                & 28357614.7              & 314692580               & 211746767               & 252115211                & 33940320.9              \\
13                    & 672523177               & 363047995               & 478726114                & 93598848.9              & 293783740               & 242667129               & 268556693                & 17400221.8              & 415836690               & 306678450               & 369831970                & 33855379.9              & 481632112               & 288395595               & 384683655                & 68455008                \\
14                    & 308380653               & 162121061               & 236215069                & 47591366.6              & 496790988               & 329700939               & 392946083                & 48819158                & 336055118               & 173434173               & 252025141                & 43405378.2              & 682758260               & 366744813               & 490368808                & 107628521               \\
15                    & 209445109               & 72629627.4              & 161148010                & 38802588.0                & 273540015               & 108530275               & 199842357                & 48473147.2              & 247701092               & 180485404               & 218347143                & 22579828.0                & 238932187               & 106568283               & 195479268                & 41676834.5              \\
\cmidrule{1-17}
16                    & 200415396               & 102244809               & 147342990                & 33477892.6              & 156055558               & 67903050                & 116403470                & 28051739.1              & 470789884               & 238359062               & 385135403                & 82014550.4              & 444685994               & 347422047               & 388895101                & 26157688.2              \\
17                    & 353293421               & 305559809               & 329706772                & 15778922.3              & 363243715               & 295126639               & 330782107                & 20721885.9              & 392264967               & 243191352               & 314440700                & 51228894.6              & 327148952               & 246849901               & 292070447                & 23650898.1              \\
18                    & 377405951               & 223747316               & 316250729                & 51127696.9              & 361215825               & 226612396               & 310344316                & 45730689.0                & 365001442               & 170015704               & 299139066                & 56015193.8              & 356645297               & 164372257               & 240253022                & 67935561.6              \\
19                    & 156685046               & 79124295.2              & 97223857.6               & 21918299.2              & 124470905               & 92660186                & 106904342                & 8835889              & 281719505               & 152039480               & 199074742                & 40268631.2              & 173314238               & 151690481               & 163403152                & 7332067.73              \\
20                    & 90566777                & 54415740.2              & 72449645.8               & 11856504.9              & 100881640               & 44903667.3              & 71143444.4               & 21069308.9              & 186475785               & 152039480               & 165831289                & 10749314.5              & 173314238               & 127021392               & 151076659                & 16539976.9\\
\bottomrule
\end{tabular}
\end{scriptsize}

\label{tab:results}
\end{sidewaystable}

We now consider the four different staging approaches with respect to NPV. In Figure~\ref{fig:whisker} we show the NPV of the remaining reserve for each time period. When looking at an individual period, the NPV contribution to the overall objective function (\ref{eq:obj}) represents a discount at time $t$. However when calculating the NPV of a mining schedule at time $t$ as done in Figure~\ref{fig:whisker}, the value of the pit is calculated by summing the per-period NPV for all future periods starting at time $t$. In the box and whisker charts, we see period $1$ with a high NPV, and it continues to grow as we remove waste material and uncover profit earning ore. As we begin to mine the ore and sell it in the later periods, the NPV decreases over time tending towards zero.
Specific to the box and whisker plots, NPV is modelled giving us a confidence range. As the deposit is mined, the confidence remains the same until large areas of uncertainty are mined out.

Figure~\ref{fig:whisker} (\subref{fig:npv_a}) shows that the lazy staging approach obtained the highest NPV value of $\$2.70$B in period $t = 4$ - very early in the life of the mine.
Although highest NPV value $\$2.60$B is achieved later in period $t = 6$ using the expected staging approach, we observe that from period $t = 12$ until the end of the mine the interquartile range of the remaining reserve is very small (see Figure~\ref{fig:whisker} (\subref{fig:npv_b})).
As expected in Figure~\ref{fig:whisker} (\subref{fig:npv_c}), the NPV value $\$3.05$B using the worst case staging approach obtained the highest value at the latest period $t = 9$.
Finally, Figure~\ref{fig:whisker} (\subref{fig:npv_d}) shows the highest NPV values $\$2.55$B for period $t = 4$ under the levelled uncertainty staging approach. We observed higher range of different NPV values until period $t = 9$. Indeed, active attempts to spread uncertainty across stages and periods seems to have inflated overall economic uncertainty. This result is counter intuitive and requires further investigation and possibly a rethink on how best to deal with geological uncertainty via staging approaches. 

Table~\ref{tab:profit_results} shows further details on the uncertainties. Although lazy staging approach achieves the highest mean profit $\$478M$ in period $t = 13$, the standard deviation is $\$94M$ which indicates the highest uncertainty among all periods. We see the same picture applying to the expected and levelled uncertainty staging approach with the highest standard deviation $(\$101M, \$108M)$ in the period
$t = 11, 14$, respectively. In contrast, the worst case staging approach attains its highest standard deviation of $\$19.5M$ in the period $t = 11$ which is only $1/5$ of maximal standard deviations of the other approaches.

\section{Discussion and Open Problems}\label{sec:discussion}
We have presented an approach for visualizing and quantifying economic uncertainty in mine planning based on geological uncertainty derived from a neural network approach to obtain multiple interpolations of the same input data. The approach allows mine planners to consider the effects of what they do not know during crucial planning periods and possibly take manual steps to mitigate downside risk.
We pointed out the impact of staging on solutions mine plans produced by Evolution.
Our results were successful in showing how different staging mechanisms can effect NPV, as well as per-period uncertainty.

However, our results do not yet inform how an algorithm could be designed to assist with this. A manual approach to staging with the intent of spreading economic uncertainty over several periods as a way of mitigating its effect on feasibility is not conclusively beneficial.
We maintain that this is a worthy goal of further research. Even if a staging approach is shown to reduce NPV, if it can conclusively and consistently be shown to mitigate the economic effects of geological uncertain using this problem model it will likely be an attractive alternative for mining operations and investors alike.

We now discuss important challenges when facing the optimisation of extraction sequences in the face of geological uncertainty in the context of using the Evolution Strategy software. We concentrate on the problem of staging. Because the optimiser used in Evolution depends on the given stage design, an important problem is to formulate a stage design that leads to a high quality schedule: in terms of maximising NPV, in terms of feasibility and now also in terms of economic mitigation of geological uncertainty. The overall goal - a remaining open problem - is to produce stage designs in an automated way that lead to feasible extraction sequences exhibiting high NPV compared with others and low economic uncertainty in critical periods of the mine plan.

From our tests we drew a number of insights. The first observation is that uncertainty in our setting can be split into two categories: domain uncertainty and grade uncertainty. Domain uncertainty typically manifests by a block switching domain between different members of the ensemble. This effect would appear to have little influence on economic uncertainty.
The impact of uncertainty observed in our experimental results is primarily related to grade uncertainty. Uncertainty in ore grade around the cut-off grade has a considerable effect on NPV. Most mines are scheduled based on their minimum cut-off grade, meaning there are typically large reserves of ore at or near the cut-off grade. If this ore was to fluctuate in grade by even a fraction of a percent, it could then be classified as waste. The effect of this re-classification is non-linear and can result in tens of millions of dollars being lost in unprofitable mining. This suggests that uncertainty around ore near the cut-off grade has a considerable effect on NPV.

Lastly, our attempts to level out uncertainty showed that in order to level out highly uncertain high grade ore, it needs to be matched with an equal amount of highly certain, high grade ore. It could also be matched with a large deposit of highly certain low to medium grade ore. This suggests that it is not just grade, but also tonnage that needs to be considered when attempting to even out uncertainty across a schedule. Further work is required here.

It would be very valuable to design automated staging approaches that lead to a feasible extraction sequence with a high NPV while minimizing the risk associated with geological uncertainty. An avenue for future work is to take the quantified economic effect of uncertainty presented here into the fitness function used as part of the evolutionary algorithm optimiser in Strategy. It could do this by processing the full ensemble simultaneously. Additional constraints around the uncertainty for critical time periods could be introduced that prevent the mine plan from having a high uncertainty around low cash flow periods where an unexpected negative cash flow could financially jeopardise the operation.

\section{Conclusions}\label{sec:conclusions}
Mine planning poses challenging and complex optimisation problems which involve high uncertainties due to the unknown grade of the ore that is mined.
In this paper, we studied the uncertainty around Maptek's mine planning software Evolution and provided a way to quantify uncertainty based on ensembles of neural networks. We have pointed out how this can be used to assess the risks associated with a given mine plan. Furthermore, we investigated the impact of staging which is provided as an input to the mine planning algorithm and have shown how staging can be used to influence the resulting solution in terms of NPV and uncertainty.

\section{Acknowledgment}
This research has been supported by the SA Government through the PRIF RCP Industry Consortium.

\bibliographystyle{abbrv}
\bibliography{references} 

\begin{thebibliography}{10}

\bibitem{DBLP:series/sci/BonyadiM16}
M.~R. Bonyadi and Z.~Michalewicz.
\newblock Evolutionary computation for real-world problems.
\newblock In S.~Matwin and J.~Mielniczuk, editors, {\em Challenges in
  Computational Statistics and Data Mining}, volume 605 of {\em Studies in
  Computational Intelligence}, pages 1--24. Springer, 2016.

\bibitem{DBLP:books/sp/2019DD}
S.~Datta and J.~P. Davim, editors.
\newblock {\em Optimization in Industry, Present Practices and Future Scopes}.
\newblock Springer, 2019.

\bibitem{DBLP:journals/jilsa/DuttaBGM10}
S.~Dutta, S.~Bandopadhyay, R.~Ganguli, and D.~Misra.
\newblock Machine learning algorithms and their application to ore reserve
  estimation of sparse and imprecise data.
\newblock {\em J. Intell. Learn. Syst. Appl.}, 2(2):86--96, 2010.

\bibitem{DBLP:journals/anor/EspinozaGMN13}
D.~G. Espinoza, M.~Goycoolea, E.~Moreno, and A.~M. Newman.
\newblock Minelib: a library of open pit mining problems.
\newblock {\em Ann. Oper. Res.}, 206(1):93--114, 2013.

\bibitem{DBLP:journals/asc/GoodfellowD16}
R.~C. Goodfellow and R.~G. Dimitrakopoulos.
\newblock Global optimization of open pit mining complexes with uncertainty.
\newblock {\em Appl. Soft Comput.}, 40:292--304, 2016.

\bibitem{DBLP:journals/jour/Ibrahimov2014}
M.~Ibrahimov, A.~Mohais, S.~Schellenberg, and Z.~Michalewicz.
\newblock Scheduling in iron ore open-pit mining.
\newblock {\em The International Journal of Advanced Manufacturing Technology},
  72(5):1021--1037, 2014.

\bibitem{DBLP:conf/Johnson69}
T.~B. Johnson.
\newblock Optimum production scheduling.
\newblock In {\em Proceedings of the 8th International Symposium on Computers
  and Operations Research, Salt Lake City}, page 539–562, 1969.

\bibitem{kennedy1990surface}
B.~A. Kennedy.
\newblock {\em Surface mining}.
\newblock Society for Mining, Metallurgy \& Exploration, 1990.

\bibitem{krige1951statistical}
D.~G. Krige.
\newblock {\em A Statistical Approach to Some Basic Mine Valuation Problems on
  the Witwatersrand}.
\newblock Chemical, Metallurgical and Mining Society of South Africa, 1951.

\bibitem{DBLP:journals/cor/LamghariD20}
A.~Lamghari and R.~G. Dimitrakopoulos.
\newblock Hyper-heuristic approaches for strategic mine planning under
  uncertainty.
\newblock {\em Comput. Oper. Res.}, 115:104590, 2020.

\bibitem{lerchs-grossmann}
H.~Lerchs and I.~F. Grossmann.
\newblock Optimum design of open-pit mines.
\newblock {\em Canadian Mining and Metallurgical Bulletin, CIM Bulletin,
  Montreal, Canada}, pages 47--54, 1965.

\bibitem{DBLP:journals/ior/LetelierEGMM20}
O.~R. Letelier, D.~G. Espinoza, M.~Goycoolea, E.~Moreno, and G.~Mu{\~{n}}oz.
\newblock Production scheduling for strategic open pit mine planning: {A}
  mixed-integer programming approach.
\newblock {\em Oper. Res.}, 68(5):1425--1444, 2020.

\bibitem{maptek}
{Maptek Pty. Ltd}.
\newblock Maptek evolution.
\newblock \url{https://www.maptek.com/products/evolution/index.html}, 2021.

\bibitem{minemax}
{MineMax Pty. Ltd}.
\newblock Minemax scheduler.
\newblock \url{https://www.minemax.com/products/scheduler/}, 2021.

\bibitem{DBLP:journals/heuristics/MontielD17}
L.~Montiel and R.~G. Dimitrakopoulos.
\newblock A heuristic approach for the stochastic optimization of mine
  production schedules.
\newblock {\em J. Heuristics}, 23(5):397--415, 2017.

\bibitem{DBLP:journals/coap/MunozEGMQL18}
G.~Mu{\~{n}}oz, D.~G. Espinoza, M.~Goycoolea, E.~Moreno, M.~Queyranne, and
  O.~R. Letelier.
\newblock A study of the bienstock-zuckerberg algorithm: applications in mining
  and resource constrained project scheduling.
\newblock {\em Comput. Optim. Appl.}, 69(2):501--534, 2018.

\bibitem{Nezamolhosseini2017}
S.~A. Nezamolhosseini, S.~H. Mojtahedzadeh, and J.~Gholamnejad.
\newblock The application of artificial neural networks to ore reserve
  estimation at choghart iron ore deposit.
\newblock {\em Analytical and Numerical Methods in Mining Engineering},
  6:73--83, 2017.

\bibitem{Ortiz2004AMA}
J.~M. Ortiz, O.~Leuangthong, and C.~V. Deutsch.
\newblock A multigaussian approach to assess block grade uncertainty.
\newblock In {\em CIM Conference and Exhibition, Edmonton 2004, May 9--12},
  pages 1--10, 2004.

\bibitem{DBLP:conf/cec/OsadaWBM13}
Y.~Osada, R.~L. While, L.~Barone, and Z.~Michalewicz.
\newblock Multi-mine planning using a multi-objective evolutionary algorithm.
\newblock In {\em {IEEE} Congress on Evolutionary Computation}, pages
  2902--2909. {IEEE}, 2013.

\bibitem{Morales2016}
J.~Saavedra-Rosas, E.~Jeivez, J.~M. Amaya, and N.~Morales.
\newblock Optimizing open-pit block scheduling with exposed ore reserve.
\newblock {\em Journal of the Southern African Institute of Mining and
  Metallurgy}, 116(7):655--662, 2016.

\bibitem{DBLP:conf/gecco/SchellenbergLM16}
S.~Schellenberg, X.~Li, and Z.~Michalewicz.
\newblock Benchmarks for the coal processing and blending problem.
\newblock In T.~Friedrich, F.~Neumann, and A.~M. Sutton, editors, {\em
  Proceedings of the 2016 on Genetic and Evolutionary Computation Conference,
  Denver, CO, USA, July 20 - 24, 2016}, pages 1005--1012. {ACM}, 2016.

\bibitem{shepard1968two}
D.~Shepard.
\newblock A two-dimensional interpolation function for irregularly-spaced data.
\newblock In {\em Proceedings of the 23rd ACM National Conference, 1968}, pages
  517--524, 1968.

\bibitem{Sullivan2019}
S.~Sullivan, C.~Green, D.~Carter, H.~Sanderson, and J.~Batchelor.
\newblock {Deep Learning - A New Paradigm for Orebody Modelling}.
\newblock {\em the AusIMM Mining Geology Conference, Perth, Western Australia
  25-26 November, 2019}, 2019.

\end{thebibliography}

\end{document}